\pdfoutput=1

\documentclass[11pt]{article}

\usepackage{ACL2023}

\usepackage{times}
\usepackage{latexsym}
\usepackage{booktabs}

\usepackage[T1]{fontenc}

\usepackage[utf8]{inputenc}

\usepackage{microtype}

\usepackage{inconsolata}

\title{Inferring Adjective Hypernyms with Language Models to Increase the Connectivity of Open English Wordnet}

\author{Lorenzo Augello\thanks{Work completed at Insight Research Ireland Centre for Data Analytics, University of Galway.} \\
  Università Cattolica del Sacro Cuore, \\
  Milan, Italy. \\
  \texttt{lorenzo.augello01@icatt.it} \And
  John P. McCrae \\
  Insight Research Ireland \\
  Centre for Data Analytics, \\
  University of Galway, \\
  Galway, Ireland. \\
  \texttt{john@mccr.ae} \\}

\begin{document}
\maketitle
\begin{abstract}
Open English Wordnet is a key resource published in OntoLex-lemon as part of the linguistic linked open data cloud. There are, however, many links missing in the resource, and in this paper, we look at how we can establish hypernymy between adjectives. We present a theoretical discussion of the hypernymy relation and how it differs for adjectives in contrast to nouns and verbs. We develop a new resource for adjective hypernymy and fine-tune large language models to predict adjective hypernymy, showing that the methodology of TaxoLLaMa can be adapted to this task.

\end{abstract}

\section{Introduction}
\label{sec:intro}

Open English Wordnet~\cite[OEWN]{mccrae-etal-2019-english} is an open-source fork of the Princeton WordNet~\cite{10.7551/mitpress/7287.001.0001}, which is modelled as and released as OntoLex data~\cite{mccrae2017ontolex}. This work aims to continue the work of maintaining the resource, as well as providing a central resource for linked data resources to connect to. However, OEWN is not itself a completely connected graph, as recent work on the verb hierarchy has shown~\cite{mccrae2025renovating}. For adjectives and adverbs, there are a very large number of synsets\footnote{`synonym set' or `synset' for short are equivalent to lexical concepts in the OntoLex model} that are not linked and this reduces the effectiveness of this resource as a central resource in Linguistic Linked Open Data (LLOD) cloud~\cite{mccrae2016open}.

Recent advances in large language models have significantly reshaped the field of lexical semantics \citep{moskvoretskii-etal-2024-large}. These models, trained on huge corpora, have demonstrated an emerging ability to infer nuanced lexical relations, and in tasks of taxonomy extraction and hypernymy discovery \citep{bordea-etal-2016-semeval}. Our goal is to use this to fully connect the graph of OEWN, however, the performance of these models has been predominantly evaluated on nouns and verbs, leaving open the question of whether they can meaningfully capture a complex relation such as hypernymy among adjectives, which is underrepresented in existing datasets and lexical resources.

Hypernymy has been extracted from large text corpora and modelled in widely used benchmarks for lexical relations classification such as BLESS \citep{baroni-lenci-2011-blessed}, EvaLution \citep{santus-etal-2015-evalution} and HyperLex \citep{vulic-etal-2017-hyperlex}. Yet, its application to adjectives remains considerably underexplored, despite its potential theoretical and practical relevance. Unlike noun and verb hypernymy, which often relies on well-defined hierarchical taxonomies, hypernymy between adjectives is inherently more fluid and context-dependent, influenced by issues such as polysemy, gradability, and contextual ambiguity \citep{Kennedy:2007}, which are often investigated by existing resources only from other relations' points of view \citep{Murphy:2003}.

Understanding hypernymy among adjectives is then critical and in order to explore its issues and despite their acknowledged complexity, our aim is to establish a clearer theory of adjective hypernymy that can offer insights into the broader organization of lexical meaning and support the development of more semantically aware language models.

In this work, we try to address the theoretical and practical gaps surrounding hypernymy between adjectives, with two main objectives. First, we propose a theoretical framework for interpreting hypernymy among adjectives and construct a gold-standard English dataset of hyponym-hypernym pairs reflecting this relation, which is released in RDF using the OntoLex model.\footnote{\url{https://github.com/lorenzoaugello/adjective-hypernymy}} Starting from existing lexical resources for other languages such as Polish WordNet \citep{maziarz-etal-2016-plwordnet} and Open Dutch WordNet \citep{postma-etal-2016-open}, we perform careful human annotation and validation to address issues of semantic ambiguity and ensure the reliability of the dataset for English by leveraging the OEWN. Second, we explore the ability of language models to recognise and interpret adjective hypernymy. We evaluate their performance both before and after training them on our gold standard, analysing how well they can capture this complex lexical relation\footnote{\url{https://huggingface.co/collections/loraug/hyper-discovery-68123c85fef0889c6559e674}}.

Examining models' performance on fine-grained and subjective semantic relations can explore and unveil their underlying linguistic competence and biases, contributing to the ongoing effort to interpret and refine their behaviour. In order for the models to be better in understanding language dynamics, we think that the development of specialized evaluation datasets is a foundational step towards benchmarking and improving their ability to disentangle complex semantic phenomena \citep{bca03b0f7d4b450094a074edaea9b6de}.

The rest of this paper is structured as follows. In Section~\ref{sec:related_work}, we discuss related work on hypernymy, adjective semantics, and language model evaluation, and in Section~\ref{sec:oewn}, we introduce the Open English Wordnet. Section~\ref{sec:theory} presents our theoretical framework, while Section~\ref{sec:dataset} describes the construction of our adjective hypernymy dataset. Section~\ref{sec:methodology} focuses on the evaluation of language models on this task, including both zero-shot testing and fine-tuning experiments. Section~\ref{sec:results} presents the results and a summary of the findings. Finally, Section~\ref{sec:conclusion} concludes the paper and suggests directions for future research, followed by the limitations of this study.

\section{Background and Related Work}
\label{sec:related_work}

Theorizing the nature of adjectives, their function, taxonomy, and interrelations has traditionally begun with categorizing how they modify nouns \citep{Raskin1995LexicalSO}, reflecting the general agreement around the definition for which adjectives serve as ``modifiers of the nouns with which they are combined'' \citep{Lyons_1977}. However, the semantic categorization of adjectives proves considerably more complex than that of nouns and verbs, as their behaviour often eludes straightforward ontological modelling \citep{mccrae-etal-2014-modelling}. While several classification attempts have been made, adjectives remain understudied in lexical semantics and lack a universal agreement on a theoretical framework. Existing classifications typically remain at a high level of general categorization, grouping adjective synsets into a few broad semantic classes known as ``supersenses'' to create taxonomies, as the first examples of \citet{Dixon+1982} and \citet{Hundsnurscher1982} show.

Moving the focus to the task of lexical entailment and the specific relation of hypernymy, we face more ambiguity. Traditional definitions of hypernymy, such as the proposal that one term A is a hypernym of another term B if A’s meaning covers the meaning of B or much broader \citep{tjong-kim-sang-2007-extracting}, are more straightforward if applied to nouns and verbs, whose hierarchical relations and structures are clearer. However, adjectives introduce challenges related to gradability, scalar structure, and context dependency \citep{Liu_Xiang_Ding_2023}. Unlike nouns, which can be organized in chains where each hyponym naturally inherits properties from its hypernyms \citep{ELX10-129}, adjectives resist simple hierarchical arrangement and prefer more compact and less vertical scales.

\noindent Notably, hypernymy relations between adjectives are absent from foundational lexical resources such as Princeton WordNet \citep{10.1145/219717.219748, 10.7551/mitpress/7287.001.0001}. Instead, adjectives are organized via antonymy (e.g., wet–dry), semantic similarity (e.g., dry–arid), and pertainymy (e.g., crime–criminal), with an absence of explicit modelling of hierarchical relations that could lead to a conflation between synonymy and hypernymy in adjectival semantics \citep{scheible-schulte-im-walde-2014-database}.

The question is now around the existence of hypernymy for adjectives. Recent works suggest that lexical relations should be conceptualized not in binary terms, but rather as graded semantic phenomena, with terms represented along a continuous scale according to principles of category membership \citep{vulic-etal-2017-hyperlex}. However, given the fact that most of the existing lexical resources and datasets created for lexical semantics tasks derive from the structure of the Princeton WordNet — which does not incorporate adjectival hypernymy — the representation and formalization of this relation remains underdeveloped.

Some lexical resources for other languages have taken steps towards modelling adjectival hypernymy. For example, the GermaNet \citep{hamp-feldweg-1997-germanet} abandons the cluster-based approach of Princeton WordNet, adopting instead a hierarchical structuring for adjectives similar to that of nouns and verbs, where, for example, ``gut'' (``good'') is the hypernym of ``toll'' (``great''). Similar hierarchical mappings are observed in the Polish WordNet, where the original ``dumbbell model'' of Princeton WordNet has been transformed into a vertical hyponymy structure \citep{rudnicka-etal-2016-challenges}. In the Open Dutch WordNet, although antonymy remains the dominant relation for adjectives, instances of hierarchical hypernymy are also introduced, such as ``knotsgek, stapelgek, krankjorum, knettergek'' (``very mad'') as hyponyms of ``gek, dwaas'' (``mad'') \citep{maks-etal-2008-adjectives}.

In the context of the NLP community, the hypernymy relation has been explored in many works, from its automatic extraction from text corpora \citep{10.3115/992133.992154} to investigations around language models' knowledge and classification capabilities \citep{ushio-etal-2021-distilling, Wang_Qiu_Huang_He_2021}. Various evaluation studies have been conducted on language models' performances, but they overwhelmingly focus on nouns \citep{camacho-collados-etal-2018-semeval}, verbs \citep{greco2024talkingtalkdoesentail}, or general natural language inference tasks \citep{madaan2024lostinferencerediscoveringrole}. Specific techniques such as prompting-based evaluation of hypernymy knowledge \citep{hanna-marecek-2021-analyzing}, dataset augmentation and benchmarking strategies \citep{kober-etal-2021-data} have been proposed, but still none of them are dedicated specifically to adjective hypernymy. Thus, while significant advances have been made in understanding lexical entailment for other parts of speech, an investigation specifically tailored to adjectives remains an unexplored field.

\section{Open English Wordnet}
\label{sec:oewn}

The \textit{Open English Wordnet} \citep{mccrae2019english} is a comprehensive, open-source lexical resource for the English language, derived from the original Princeton WordNet \citep{10.7551/mitpress/7287.001.0001}, and developed using the W3C OntoLex-Lemon model \citep{mccrae2017ontolex}. As an OntoLex resource, OEWN publishes lexical data on the Web in accordance with linked data principles, thereby facilitating interoperability across linguistic and semantic resources.

OEWN plays a central role in the Linguistic Linked Open Data (LLOD) cloud \citep{mccrae2016open}, acting as a foundational resource for representing lexical semantics, enabling cross-linguistic comparison, and linking to other language resources. It uses the OntoLex-Lemon model as a standardized vocabulary for representing lexical information as RDF. The Global WordNet Association has developed formats for the representation of wordnets~\cite{mccrae2021globalwordnet}, which extend the core principles of OntoLex. These formats support multiple serializations in XML, JSON(-LD), and RDF/Turtle—providing interoperability with other OntoLex resources. These formats have been adopted by OEWN and other wordnets in the Open Multilingual Wordnet (OMW) to ensure compatibility and extensibility.

In RDF, the GWA format directly encodes \texttt{LexicalEntry}, \texttt{LexicalSense}, and \texttt{LexicalConcept} using OntoLex classes, while WordNet-specific metadata and relations (e.g., sense keys, synset ordering) are modelled via a dedicated ontology at \url{https://globalwordnet.github.io/schemas/wn}.

\section{Theoretical proposal}
\label{sec:theory}

Even though it is true that hypernymy for adjectives is difficult to define and has unclear boundaries, we think that it is a necessary relation that could help in the semantic organization of a language by introducing a hierarchy of meaning and usage that is not limited to a simple categorization in broad classes or to fuzzy and often confused horizontal and opposition relations. The distinction with synonymy is in fact very narrow, but we think that in the scope of the same semantic field, there are adjectives with a broader meaning than others, and those need to be distinguished (e.g., cognitive - mental - immaterial; displeasing - unpleasant - negative). 

The Princeton WordNet has 117,659 synsets and 84,428 hypernym-hyponym relations, but those are only for nouns and verbs, and not for adjectives and adverbs. Furthermore, there is currently no reliable dataset for the English language that represents this relation for adjectives. Facing the limited availability of high-quality training data, one of the main objectives of this study was thus the creation of a reliable initial dataset containing adjective pairs and their definitions, drawn from the OEWN, in order to address the problem of word-sense disambiguation.

Throughout the annotation and reliability evaluation of hyponym-hypernym pairs extracted from existing lexical resources, a principle of substitution was followed (see \citet{mccarthy-navigli-2007-semeval} for other applications of lexical substitution): the hypernym should be substitutable in place of the hyponym in context, such that the resulting sentence preserves the original meaning at an acceptable general level, without contradiction or need for forced interpretation. While a semantic broadening is both expected and required, the sense of the hyponym needs to be included within the possible interpretations of the hypernym without ambiguity. Thus, hypernymy is characterized by inclusion of meaning, but not equivalence, distinguishing itself from synonymy. In the annotation process, if the connection between the two adjectives of the candidate pair was not perceived as direct and intuitive by the annotators, the pair was discarded, as we do not want dubious inference or multiple plausible readings. Given that hypernymy is inherently difficult to define - being vague and often subject to flexible and debatable boundaries - only those pairs deemed most reliable were retained (even though, undoubtedly, some may still be debated, highlighting once again the subjective nature of such a relation).

One of the main challenges in identifying a hypernymy relation between two adjectives lies in the polysemous nature of many of them. A first example is “cold”, which can refer either to the semantic field of temperature or to aspects of human behaviour and personality. Another case is “hard”, which in one sense refers to a physically solid material, and in another to the more abstract concept of difficulty. Adjectives' meaning and their relations are therefore defined not only in isolation but, more importantly, depending on the associated noun and the semantic domain they evoke: “cold” can describe both an environment and a person; “hard” can qualify both a material and a problem. In this regard, context plays a crucial role. While this is already a challenge for human annotators, it becomes even more complex for language models that have to deal with semantic disambiguation (see Section~\ref{sec:dataset}). Moreover, many English words can assume more than one part of speech depending on context and usage, without undergoing inflectional changes, as English has a very limited morphological richness. A first example is the word “clean”, which can function as both a verb and an adjective. A broader and more frequent case involves present and past participles, which are at times interpreted as verbs and at other times as adjectives. This ambiguity can be extended to nouns as well (the above mentioned word ``cold'' can have multiple adjectival meanings, but it can also be a noun: ``a mild viral infection involving the nose and respiratory passages''), and given the pervasiveness of this phenomenon in English, it was deemed necessary to incorporate OEWN definitions for both hyponyms and hypernyms into the construction of the gold standard dataset (see also the experiments of \citet{moskvoretskii-etal-2024-large} for why definitions are crucial).

In contrast to the more straightforward case of nominal hypernymy, we think that adjectival hypernymy needs to be grounded in sense disambiguation, operationalized through substitution-based inclusion tests, and evaluated in a context. Our gold-standard dataset leverages those principles and aims to provide an initial but reliable resource for further studies of adjectives' semantics and its modelling in computational systems.

\section{Dataset creation}
\label{sec:dataset}

The construction of the gold-standard dataset was based on pre-existing lexical resources, leveraging wordnets from languages that explicitly encode adjectival hypernymy relations: the Open Dutch WordNet (ODWN) and the Polish WordNet (plWN). These two resources were chosen because, unlike the Princeton WordNet, they organize adjectives hierarchically, making them suitable starting points for our purposes.

Hyponym-hypernym pairs of adjectives were automatically extracted from those two resources by using the \texttt{wn} Python library\footnote{\url{https://pypi.org/project/wn/}}, and an initial pool of 450 pairs was randomly picked from the total (166 from ODWN and 284 from plWN). Only adjectives were considered, and both child hyponym and father hypernym nodes were included. Each extracted pair was then manually translated into English using bilingual online dictionaries (Wiktionary and the Cambridge Dictionary), consistently selecting the first suggested translation to minimize subjective bias and arbitrary choice. 

After this, each pair was reviewed individually by two annotators. If both adjectives of the pair appeared in the same synset within the OEWN - indicating synonymy rather than a hierarchical relation - the pair was discarded and the original relation was kept: for example, ``difficult'' and ``hard'', which are hypernym-hyponym in plWN, were not included because they are synonyms in OEWN (“not easy; requiring great physical or mental effort to accomplish or comprehend or endure”). When multiple hypernyms were available for a hyponym, if they belonged to the same OEWN synset they were kept; otherwise, only the most reliable one was chosen and agreed upon by the annotators, while the others were discarded.

So, starting from the initial 450 pairs, 148 were discarded, and the final gold standard for English comprises 302 adjective pairs: 92 sourced from ODWN, 170 from plWN, and 40 derived from either of the two (18 ODWN, 22 plWN) but for which a more reliable alternative hypernym was proposed and approved by the annotators (e.g., the plWN had ``effective'' as hypernym for ``deft'', but ``skillful'' was suggested and approved as an alternative and kept in the dataset). Acceptance rates were similar across resources, with a retention ratio of 0.55 for ODWN (92/166) and 0.60 for plWN (170/284).
Furthermore, as already mentioned in Section~\ref{sec:theory}, for each adjective included in the gold standard, the corresponding English definition was included after retrieving it from the OEWN. This was essential to ensure that the annotated relations reflected the intended senses of the adjectives rather than possible polysemous variants.

The annotation process began with an initial small dataset of 50 adjective pairs, annotated by two annotators using a three-label classification system (yes, no, maybe). All cases of negative agreement (no-no), strong disagreement (yes–no), weaker disagreement (no–maybe) and shared doubt (maybe-maybe) were discarded, while complete agreement (yes–yes) cases were retained and partial agreement (yes–maybe) ones were further discussed (examples are shown in Table~\ref{table:1}). From this first sample, 62\% (31 adjective pairs) were selected (Cohen's kappa $\kappa=0.65$). From a second sample of 100 pairs, 69\% were retained ($\kappa=0.64$), and from a third sample of 300 pairs, 67\% (201 pairs) were selected ($\kappa=0.61$). A third annotator was then included for the annotation of 100 pairs randomly selected from the third larger sample (300 pairs), providing additional discussion and validation (Fleiss' kappa $\kappa=0.48$).
\begin{table}
\small\centering
\begin{tabular}{lll}
\hline
\textbf{Annotation} & \textbf{hypo} & \textbf{hyper}\\
\hline
{yes-no} & {intelligent} & {rational}\\
{yes-maybe} & {limitless} & {vast}\\
{yes-yes} & {lucid} & {aware}\\
{no-maybe} & {multiple} & {plural}\\ 
{maybe-maybe} & {unneeded} & {useless}\\ 
{no-no} & {productive} & {rich}\\
\hline
\end{tabular}
\caption{Examples of agreements and disagreements in the annotation of the pairs of adjectives performed by the two annotators.}
\label{table:1}
\end{table}

Given the difficulty of semantic relations tasks which introduce a high level of subjectivity, inter-annotator agreement levels were moderate, but still acceptable, consistent and comparable across the different samples, with most disagreements involving ``maybe'' labels rather than direct contradictions (15 occurrences of the ``maybe'' label in the first sample, 24 in the second and 93 in the third). 

In addition to the main gold standard where each hyponym is associated to one single exact hypernym, a second version of the dataset was later developed for model fine-tuning and evaluation (see Section~\ref{subsec:traindetails}). Here, synonyms of each hypernym were added based on synset membership in OEWN, in order to account for cases where multiple semantically correct hypernyms exist (see Table~\ref{table:2}). This was motivated both by the natural multiplicity of hypernyms for a given adjective and for reducing the possible influence of definition-based prompts during the evaluation of the models.

\begin{table*}[!ht]
\small\centering
\begin{tabular}{lllp{25mm}p{25mm}}
\hline
\textbf{Dataset} & \textbf{hyponym-lemma} & \textbf{hypo\_definition} & \textbf{hypernym-lemma} & \textbf{hyper\_definition}\\
\hline
{single} & {relaxed} & {without strain or anxiety} & {calm} & {not agitated; without losing self-possession}\\
{multiple} & {relaxed} & {without strain or anxiety} & {calm, serene, tranquil, unagitated} & {not agitated; without losing self-possession}\\
\hline
\end{tabular}
\caption{One example of a pair taken from the two versions of the gold standard, first showing one single exact hypernym (``calm'') for the input hyponym (``relaxed''), and then showing its synonyms found in OEWN too.}
\label{table:2}
\end{table*}

\section{Methodology}
\label{sec:methodology}

From the definition of hypernymy for adjectives to the issues in creating a benchmark dataset, it is clear that the task of lexical entailment already presents many issues for human understanding. As for the capabilities of language models, there have been attempts to assess their reliability following either a binary classification approach (choosing the correct hypernym) or a generation one (given a hyponym, predicting its most probable hypernym). 

In this study, we use two models: TaxoLLaMa and SmolLM-360M-Instruct. We first test their capabilities in the hypernymy discovery task and then fine-tune them on both our benchmark datasets in order to explore their capabilities of capturing semantic knowledge from them.

TaxoLLaMa \citep{Moskvoretskii_2024} is the finetuned version of the LLaMA-2-7b model \citep{touvron2023llama2openfoundation}, which was trained on the WordNet dataset for 16 taxonomy-related tasks and reached SoTA results on hypernymy discovery in different domains and languages. As it was neither trained nor tested on adjective examples, however, its performance in predicting them was expected to be lower.

SmolLM-360M-Instruct is part of SmolLM, a family of state-of-the-art small models\footnote{\url{https://huggingface.co/blog/smollm}}. It is optimized for instruction-following tasks through supervised fine-tuning on multiple instruction datasets. Even though it is much smaller than TaxoLLaMa, its size makes it ideal for rapid fine-tuning and evaluation on specialized tasks such as adjective hypernymy discovery, enabling us to explore if smaller models can acquire semantic relations when provided with limited data. 

\subsection{Training Details}
\label{subsec:traindetails}

We performed the fine-tuning using the Unsloth method\footnote{\url{https://unsloth.ai/}}, which allowed us to quantize the models to 4
bits and train them using LoRA \citep{hu2022lora}, reducing memory and computational requirements without affecting performance. This approach was particularly suited for using our small datasets and enabled the fine-tuning of both smaller and larger models like TaxoLLaMa with limited hardware resources and easy accessibility\footnote{\url{https://colab.research.google.com/}}. We fine-tuned the models using the SFTTrainer class, conducting the training for 60 optimization steps with a learning rate of 2e-4. The per-device batch size was set to 2, and gradient accumulation was used with 4 steps, simulating a batch size of 8. Leveraging our gold standard dataset, we fine-tuned the models on a training set composed of 211 items (70\% of the dataset). The one below is a training sample, with an input question by the human user and an expected output answer by the GPT assistant, following a chat template: 

\begin{itemize} 
\item[] (from: human) ``What are the hypernyms of the hyponym: ``complicated'' (definition: ``difficult to analyze or understand'')?'',
\item[] (from: gpt) ``The hypernyms are: ``difficult, hard'' (definition: ``not easy; requiring great physical or mental effort to accomplish or comprehend or endure'').''
\end{itemize}

The training was performed in two separate settings: first the models were fine-tuned using the original gold standard dataset (hereafter referred to as ``single'') with a one-to-one correspondence between each hyponym and its relative hypernym; then a different gold standard was developed (``multiple''), implementing the single dataset with the synonyms of the hypernyms in order to have a one-to-many correspondence between each hyponym and its relative hypernyms (see Table~\ref{table:2}). In order to have a consistent criterion and not fall into ambiguities, we relied on the OEWN and added all the adjectives pertaining to the same synsets of the original single hypernyms. This was done for two main reasons. First, given that a hyponym can have more than one hypernym, including only one correct exact hypernym in the gold standard could leave out other possible candidates (e.g., ``opportune'' has ``suitable'' as its exact hypernym in the single dataset, but ``appropriate'' and ``suited'' were added in the multiple dataset, pertaining to the same synset: ``meant or adapted for an occasion or use''). Secondly, the models trained on the single dataset would output only one hypernym for each hyponym when tested, while if trained with multiple possible hypernyms they would include more predictions and improve their semantic knowledge.

\subsection{Prompting}

Both during the zero-shot evaluation and after the fine-tuning, we used two different prompt settings: first, we just provided the models with an input hyponym adjective, and then we also gave them the hyponym definition. For the zero-shot prompting, we followed the below format originally used for TaxoLLaMa:

\begin{itemize} 
\item[] <s>[INST] <<SYS>> You are a helpful assistant. List all the possible words divided with a comma. Your answer should not include anything except the words divided by a comma<</SYS>>
\item[] hyponym: humorous (full of or characterized by humor) | hypernyms: [/INST]
\end{itemize}

\noindent While after the fine-tuning we followed the below format: 

\begin{itemize} 
\item[] messages = [{``from'': ``human'', ``value'': ``What are the hypernyms of the hyponym: ``invigorating'' (definition: ``imparting strength and vitality'')?''},]
\end{itemize}

\section{Results}
\label{sec:results}

The original base TaxoLLaMa model reached SoTA results and scored an MRR (Mean Reciprocal Rank) of 54.39 for English, but this was trained only on verbs and nouns sampled from the WordNet-3.0 graph. So, its performance on hypernymy discovery for adjectives was expected to be lower. In order to assess this, we test it in a zero-shot setting on our test set (91 pairs, 30\% of the total gold standard), and we record an MRR of 9.4 when the model is not prompted with the definition of the input hyponym, and an increase to 25.8 when the definition is given. After the fine-tuning on the multiple dataset with the synonyms, those scores improve, respectively, to 23.6 and 33.3. 

The difficulty in treating adjectives and distinguishing them from other parts of speech was a major one with the base TaxoLLaMa: when evaluated on the test set in a zero-shot setting without the definition, 58\% of times it gives as output only nouns, 21\% only adjectives, and 21\% both (by providing it with the definition, the numbers change respectively to 14, 44 and 42\%). After the fine-tuning on the single dataset, the amount of predicted adjectives significantly increases: 95\% without the definition and 100\% with the definition (both 100\% when fine-tuned on the multiple dataset). This improvement is achieved by both TaxoLLaMa and SmolLM-360M-Instruct, as Table~\ref{table:3} shows. Given the difficulty with POS recognition and disambiguation, we consider this result to be almost as relevant and significant as the correct hypernym prediction.  

\begin{table}[h!]
\small\centering
\begin{tabular}{llcc}
\hline
{Model} &  {Setting} & {No def} & {With def}\\
\hline
\verb|TaxoLLaMa| & {Zero-shot} & {0.39} & {0.78}\\
\verb|TaxoLLaMa| & {ft-single} & {0.96} & {1.00}\\
\verb|TaxoLLaMa| & {ft-multi} & {1.00} & {1.00}\\
\verb|SmolLM-360M-Instr| & {Zero-shot} & {0.69} & {0.77}\\ 
\verb|SmolLM-360M-Instr| & {ft-single} & {0.79} & {0.96}\\ 
\verb|SmolLM-360M-Instr| & {ft-multi} & {1.00} & {1.00}\\ 
\hline
\end{tabular}
\caption{F1-score performances on predicting the correct POS (ADJ), before the fine-tuning (Zero-shot), when trained on the single dataset (ft-single) and when trained on the multiple dataset (ft-multi), without and with the definition.}
\label{table:3}
\end{table}

After the fine-tuning, the models were tested in two settings against the two different datasets, in order to first evaluate their capability of inferring the exact correct hypernym of a given hyponym (against the single dataset), and then introducing also the possibility of giving synonyms in output (against the multiple dataset). 

Table~\ref{table:4} shows the results in the first setting, where TaxoLLaMa-ft-multi reaches the best scores both without and with the definition in the prompt. Interestingly, providing the definition does not improve the performance of TaxoLLaMa-ft-single. At first glance, this may seem surprising. However, we need to consider that this model produces only one adjective as output, and this choice is often strongly influenced by lexical overlap with words in the input definition: e.g., ``extant'' (defined as ``still in existence; not extinct or destroyed or lost'') has ``real'' as its correct hypernym, but TaxoLLaMa-ft-single predicts ``existent'', being biased by the vocabulary of the definition. 

\begin{table}[h!]
\small\centering
\begin{tabular}{llcc}
\hline
{Model} & {Setting} & {No def} & {With def}\\
\hline
\verb|TaxoLLaMa| & Zero-shot & {0.15} & {0.39}\\
\verb|TaxoLLaMa| & ft-single & {0.32} & {0.31}\\
\verb|TaxoLLaMa| & ft-multi & {0.35} & {0.44}\\
\verb|SmolLM-360M-Instr| & Zero-shot & {0.13} & {0.16}\\ 
\verb|SmolLM-360M-Instr| & ft-single & {0.14} & {0.15}\\ 
\verb|SmolLM-360M-Instr| & ft-multi & {0.21} & {0.25}\\
\hline
\end{tabular}
\caption{Performances evaluated against the single dataset on predicting the exact correct hypernym before the fine-tuning, when trained on the single dataset and when trained on the multiple dataset, without and with the definition. Given that apart from TaxoLLaMa-Zero-shot all the other models output a single hypernym, precision and recall are equal and only one value is reported.}
\label{table:4}
\end{table}

\begin{table*}[]
    \small\centering
    \begin{tabular}{p{30mm}p{20mm}|ccc|ccc}
    \hline
    Model & Setting & \multicolumn{3}{c|}{No def} & \multicolumn{3}{c}{With def}\\
     & & P & R & F-M & P & R & F-M \\
     \hline
     \verb|TaxoLLaMa| & Zero-shot & {0.04} & {0.13} & {0.06} & {0.07} & {0.23} & {0.11}\\
    \verb|TaxoLLaMa| & ft-single & {0.14} & {0.14} & {0.14} & {0.16} & {0.16} & {0.16}\\
    \verb|TaxoLLaMa| & ft-multi & {0.15} & {0.20} & {0.17} & {0.28} & {0.25} & {0.26}\\
    \verb|SmolLM-360M-Instr| & Zero-shot & {0.07} & {0.07} & {0.07} & {0.09} & {0.09} & {0.09}\\ 
    \verb|SmolLM-360M-Instr| & ft-single & {0.09} & {0.09} & {0.09} & {0.10} & {0.10} & {0.10}\\ 
    \verb|SmolLM-360M-Instr| & ft-multi & {0.16} & {0.10} & {0.12} & {0.20} & {0.14} & {0.16}\\
     
     \hline
    \end{tabular}
    \caption{Performances evaluated against the multiple dataset on predicting a list of possible hypernyms before the fine-tuning, when trained on the single dataset and when trained on the multiple dataset, without and with the definition. For the models that output only one hypernym (TaxoLLama-ft-single, SmolLM-Zero-shot and SmolLM-ft-single, precision, recall and F-measure values are the same.}
    \label{table:5}
\end{table*}

This variability and influence due to definitions was also one of the motivations to introduce synonyms into the multiple gold standard (``real'' and ``existent'' are in the same synset in OEWN). By allowing synonyms, we can consider acceptable also cases where the model predicts a semantically correct hypernym which is not exactly the one we expected. Additionally, it is important to note that the base TaxoLLaMa is trained to output a list of hypernyms, whereas TaxoLLaMa-ft-single produces only a single prediction. This structural difference increases the chances of including a correct hypernym in the output for the base model (resulting in a higher F1-score of 0.39 against 0.31). However, when evaluating only the first-ranked hypernym from the base TaxoLLaMa’s output list, its performance drops, with only 7 out of 14 correct hypernyms appearing in the first position.

The results shown in Table~\ref{table:5}, obtained in the second evaluation setting against the multiple dataset, reveal that TaxoLLaMa-ft-multi performs better than the other models, and we observe a consistent improvement in the performance of the models when definitions are included in the prompt, unveiling the importance of incorporating them for disambiguation. Additionally, when compared to their counterparts trained on the single dataset (TaxoLLaMa-ft-single and SmolLM-ft-single), both TaxoLLaMa-ft-multi and SmolLM-ft-multi demonstrate clear gains, highlighting the benefit of allowing multiple valid hypernyms during training for capturing the nuanced nature of the hypernymy relation. 

\section{Conclusion}
\label{sec:conclusion}

In this paper, we explore the understudied relation of adjective hypernymy, both from a theoretical and a computational perspective. We propose a definition for it, grounded in semantic inclusion and contextual substitutability, distinguishing it from synonymy and other relations. Making use of this framework, we construct a two-version gold-standard English dataset for adjective hypernymy by adapting the lexical information stored in the Polish, Dutch and Open English wordnets. Our dataset was validated through human annotation and synset-based disambiguation, offering a reliable, small and initial benchmark for future research.

We then evaluate the capabilities of language models - the large-scale TaxoLLaMa and the smaller SmolLM-360M-Instruct - on the task of hypernymy discovery, both in zero-shot and fine-tuned settings. Our results show that the models initially struggle with adjective hypernymy, particularly with issues of POS ambiguity and semantic polysemy. However, after fine-tuning on our dataset, especially the synonym-augmented variant, their performances improve, highlighting the value of task-specific training data. We also found that providing the models with explicit definitions of the input adjectives improves their ability to identify correct hypernyms. This underscores the central role of word sense disambiguation in first identifying and then modelling adjectival meaning. 

As future work, we would like to a) expand the gold-standard dataset to reach a higher coverage and more generalizability, and allow for a better theoretical description and more reliability in training and evaluating language models, b) model adjective hypernymy in OEWN, identifying and representing hypernymy relations for all adjectives in the resource, c) support the interoperability between the OntoLex and NLP communities, promoting the use of adjective hypernymy in downstream applications, d) extend language models evaluation to other semantic relations between adjectives to explore how they differ theoretically and how well they are captured computationally.

\section*{Limitations}

\begin{enumerate}
    \item Dataset size: The gold standard is limited in size, which may restrict both its theoretical completeness (in modelling the full spectrum of adjectival hypernymy) and its practical utility (for model training and evaluation, especially on unseen or ambiguous adjective pairs), constraining generalizability and lexical coverage.
    \item Models evaluated: Only two language models were used, so this limits the conclusions regarding models capabilities by necessarily leaving out other existing architectures, sizes, and training data diversities.
    \item Subjectivity of annotation: The annotation process, although based on carefully defined criteria and multi-annotator agreement, introduces a degree of subjectivity, which is even more accentuated by the graded, nuanced and polysemous nature of adjective hypernymy itself.
    \item Language: The dataset, the theories and the evaluation are all limited to English. Consequently, some of the observed phenomena may not be generalized cross-linguistically, as the semantic behaviour of English adjectives may be different from those of other languages.

\end{enumerate}

\bibliographystyle{acl_natbib}
\bibliography{anthology, references, publications}

\end{document}